\documentclass[letterpaper, 10pt, conference]{ieeeconf}      

\IEEEoverridecommandlockouts                              
\usepackage{amsmath} 
\usepackage{amssymb}  
\usepackage{cite}
\usepackage{threeparttable}
\usepackage{stfloats}
\usepackage{comment}
\usepackage{gensymb}
\usepackage{subfig}
\usepackage{booktabs} 
\usepackage{hyperref}

\title{\LARGE \bf
COCOI: Contact-aware Online Context Inference for \\
Generalizable Non-planar Pushing
}

\author{
       Zhuo Xu$^{1, 2*}$,
       Wenhao Yu$^{3}$, 
       Alexander Herzog$^1$,
       Wenlong Lu$^{1}$, 
       Chuyuan Fu$^{1}$,\\
       Masayoshi Tomizuka$^{2}$,
       Yunfei Bai$^{1}$, 
       C. Karen Liu$^{4}$,
       Daniel Ho$^{1}$

\thanks{$^{1}$Everyday Robots, X The Moonshot Factory, Mountain View, CA, USA}%
\thanks{$^{2}$University of California, Berkeley, Berkeley, CA, USA}%
\thanks{$^{3}$Robotics at Google, Mountain View, CA, USA}%
\thanks{$^{4}$Stanford University, Stanford, CA, USA}%
\thanks{$^{*}$Work done as an AI Resident at Everyday Robots.}
}

\usepackage{algorithm,algpseudocode}
\usepackage[algo2e]{algorithm2e} 
\usepackage{graphicx}\usepackage{multirow}
\usepackage{url}
\usepackage{color,soul}
\usepackage{bm}

 \newcommand{\yunfei}[1]{\textcolor{red}{}}
 \newcommand{\wenhao}[1]{\textcolor{cyan}{}}
 \newcommand{\karen}[1]{\textcolor{blue}{}}
 \definecolor{mypink}{rgb}{0.858, 0.188, 0.478}
\newcommand{\zhuoxu}[1]{}
\begin{document}

\maketitle
\thispagestyle{empty}
\pagestyle{empty}

\begin{abstract}
General contact-rich manipulation problems are long-standing challenges in robotics due to the difficulty of understanding complicated contact physics. Deep reinforcement learning (RL) has shown great potential in solving robot manipulation tasks. However, existing RL policies have limited adaptability to environments with diverse dynamics properties, which is pivotal in solving many contact-rich manipulation tasks. In this work, we propose Contact-aware Online COntext Inference (COCOI), a deep RL method that encodes a context embedding of dynamics properties online using contact-rich interactions. We study this method based on a novel and challenging non-planar pushing task, where the robot uses a monocular camera image and wrist force torque sensor reading to push an object to a goal location while keeping it upright. We run extensive experiments to demonstrate the capability of COCOI in a wide range of settings and dynamics properties in simulation, and also in a sim-to-real transfer scenario on a real robot (Video: \url{https://youtu.be/nrmJYksh1Kc}).

\end{abstract}

\section{Introduction}

Contact rich manipulation problems are ubiquitous in the physical world. In millions of years of evolution, humans have developed the remarkable capability to understand environment physics, so as to achieve general contact rich manipulation skills. Combining visual and tactile perception with end-effectors like fingers and palms, humans effortlessly manipulate objects with various shapes and dynamics properties in complex environments. Robots, on the other hand, lack this capability -- due to the difficulty of understanding high dimensional perception and complicated contact physics. Recent development in deep reinforcement learning (RL) has shown great potential towards solving manipulation problems \cite{levine2016end, kalashnikov2018qt, lee2020learning} by leveraging two key advantages. First, the representative capability of a deep neural network structure can capture complicated dynamics models. Second, control policy optimization explores vast contact interactions. However, contact-rich manipulation tasks are generally dynamics-dependent; since the RL policies are trained in a specific dynamics setting, they specialize within the training scenario and are vulnerable to variations of dynamics. Learning a policy that is robust to dynamics variations is pivotal for deployment to scenarios with diverse object dynamics properties.

\begin{figure}
\centering
\includegraphics[width=0.9\linewidth]{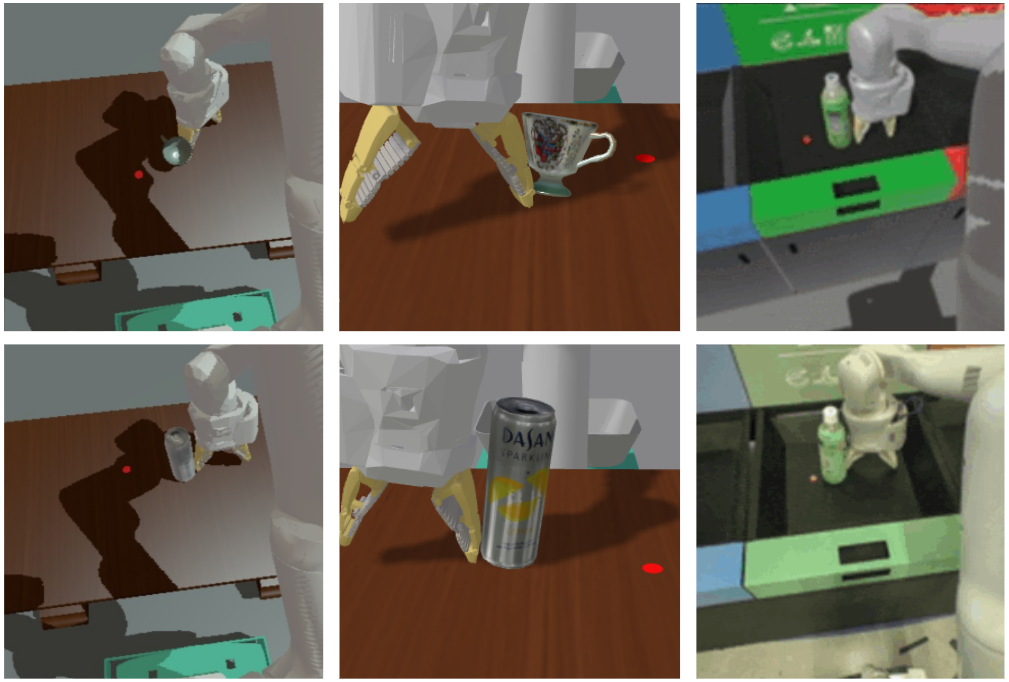}
\caption{Our method, COCOI, achieves dynamic-aware, non-planar pushing of an upright 3D object. The method is robust against domain variations, including various objects and environments, in both simulation and the real world. The first and second columns show the table simulation setting in the robot's perspective and the third party perspective, respectively. The third column shows the simulated and real world trash bin settings in the robot perspective.}
\label{fig:overview}
\end{figure}

In this work, we design a deep RL method that takes multi-modal perception input and uses deep representative structure to capture contact-rich dynamics properties. The proposed method, Contact-aware Online COntext Inference (COCOI), uses prior camera frames and force readings in a contact-aware way to encode dynamics information into a latent context representation. This allows the RL policy to plan with dynamics-awareness and improves in robustness against domain variations. 

We apply COCOI to a novel pushing task where dynamics property reasoning plays a vital role: the robot needs to push an object to a target location while avoiding knocking it over (Fig. \ref{fig:overview}). Prior work in pushing mostly focus on objects inherently stable when pushed on a flat surface. This essentially reduces the task to a 2D planar problem \cite{zhou2016convex} \cite{li2018push}. As a result, they cannot handle our proposed class of ``non-planar pushing tasks'' where real-world 3D objects can move with the full six degrees of freedom during pushing. Despite being commonly seen in everyday life, these tasks have the following challenges: 

\begin{enumerate}
\item Visual perception: unlike in planar pushing, where concrete features can be retrieved from a top down view, in non-planar pushing, key information can not be easily extracted from the third angle perspective image.
\item Contact-rich dynamics: the task dynamics properties are not directly observable from raw sensor information. Furthermore, in our non-planar pushing task, dynamics property reasoning is vital to avoid knocking the object over. 
\item Generalizable across domain variations: the policy needs to be effective for objects with different appearances, shapes, masses, and friction properties. 
\end{enumerate}


The paper is structured as follows: In Section II, we review and compare with related works. We formalize the problem of interest in Section III and describe the RL pushing controller in Section IV. We explain COCOI in Section V and the handling of the sim-to-real visual gap in Section VI. The capability of COCOI is validated with the experiments in Section VII, before we make the conclusions in Section VIII.
\section{Related Work}

In order to adapt a learning based policy to different task settings, variation of dynamics, and unknown disturbances, previous researchers train policies on randomized environments with possible variations to gain robustness. Rajeswaran et al. \cite{rajeswaran2016epopt}. and Peng et al. \cite{peng2018sim} learn RL policies in randomized simulated environments and directly apply them to different domains. Chebotar et al. leverage real world experiences to adapt the simulation randomization \cite{chebotar2019closing}. Other work aim to derive a representation of the task context for domain specific planning. Rakelly et al. take a meta learning direction to learn patterns within the task to help plan \cite{rakelly2019efficient}. Yu et al. learn a universal policy and use an online system identification module to learn the dynamics parameters \cite{yu2017preparing}. Beyond learning based methods, planning-control framework can also overcome the domain gap. Harrison et al. use model predictive control to track a reference trajectory derived from a learning policy \cite{harrison2020adapt}, and Xu et al. use a robust controller to reject dynamics variation and external disturbances \cite{xu2018zero}\cite{tang2019disturbance}. COCOI is different from previous methods in that we focus on the contact-rich manipulation problem, and use a contact-aware structure to infer the contact dynamics context online.

Another thread of related research is the pushing task \cite{stuber2020let}. Earlier studies on pushing are based on analytical approaches and consider quasi-static planar pushing \cite{mason1986mechanics}. Later, researchers have introduced data driven methods to model pushing physics. Zhou et al. develop a dynamics model for planar friction and design force control method for planar sliding \cite{zhou2016convex}. Yu, Bauza et al created a planar pushing dataset for data-driven modeling \cite{yu2016more, bauza2017probabilistic}. More recent works involve using deep learning to learn object properties \cite{li2018push,ajay2018augmenting, xu2019densephysnet}, but they only focus on planar pushing problems. Stuber et al. \cite{stuber2018feature} and Byravan et al. \cite{byravan2017se3} learn motion models for pushing simple blocks.

On non-planar pushing, the majority of works are still in the exploration stage. Ridge et al. propose an object image fragment matching method for 3D objects \cite{ridge2015self}, albeit for a limited library of objects. Zhu et al. use a simulator as a predictive model \cite{zhu2017information}. Kopicki et al. learn a 3D dynamics model in the PhysX simulator \cite{kopicki2017learning}. These works all rely on carefully selected objects and precise detection and localization; our model takes monocular camera image input and solves a generalized non-planar pushing task for diverse objects.

\section{Problem Statement}
In this section, we formally define the proposed non-planar pushing task and formulate the problem using a Partially Observable Markov Decision Process (POMDP). We focus on the class of pushing tasks where maintaining the upright pose of the object is critical. For example, when pushing a glass of water on the table, the glass should not tilt and spill. In our work, we assume that an object is randomly placed on a flat surface with an upright initial pose. The task objective is to use the robot end effector to push the object to a random target position on the surface, while maintaining its upright orientation. The object can have irregular shape and mass distribution, and the robot may push at any point on the object, making the contact dynamics physics complicated.

We use state $\bm{s}\in S$ to represent the full task state. In a POMDP, the state is not directly available and can only be inferred using observations. Concretely, we use the image captured using the robot's monocular camera as the high dimensional observation (Fig. \ref{fig:overview}, first and third columns), in which the target location is rendered using a red dot. We also include a low dimensional proprioception observation of the gripper height and open/close status. We use state and observation interchangeably in the following sections. The action $\bm{a} \in A$ contains the designated position and orientation of the gripper, the gripper open/close command, and a termination boolean. The system transition function $T: S \times A \rightarrow S$ follows the physical model, and we use a sparse binary reward function $R: S \times A \rightarrow \mathbb{R}$ with value 1 when the distance between the object and the target is smaller than a threshold and the object is upright. The task objective is to maximize the expectation of the discounted cumulative reward, or the return 
\begin{equation}
R = \mathop{\mathbb{E}}_{\{(\bm{s_t}, \bm{a_t})\}} \left[\sum^{t = \infty}_{t = 0} \gamma^{t} r_t\right]
\end{equation}
where $\gamma$ is the discount coefficient, and $r_t = R(\bm{s_t}, \bm{a_t})$ is the reward obtained at time $t$.

\section{Learning Basic Pushing Controller}
\label{baseline}

We use Q-learning to train object pushing control policy. The definition of the Q function is
\begin{equation}
Q(\bm{s}, \bm{a}) = \mathop{\mathbb{E}}_{\{(\bm{s_t}, \bm{a_t})\}} \left[\sum^{t = \infty}_{\bm{s}, \bm{a}} \gamma^{t} r_t\right],
\end{equation}
the expected return starting from state $\bm{s}$ and taking $\bm{a}$. The Q function satisfies the Bellman function:
\begin{equation}
Q(\bm{s_t}, \bm{a_t}) = \mathbb{E}_{\bm{s_{t+1}}} [R(\bm{s_t}, \bm{a_t}) + \gamma \cdot Q(\bm{s_{t+1}}, \pi(\bm{s_{t+1}}))]
\end{equation}
where $\pi(\bm{s_{t+1}})$ represents the action selected by the optimal policy at the current iteration and state $\bm{s_{t+1}}$. The optimal policy is defined by:
\begin{equation}
\pi(\bm{s_t}) = \arg\max_{\bm{a} \in A} Q(\bm{s_t}, \bm{a})
\end{equation}
One Q-learning iteration uses the collected data tuples $(\bm{s_t}, \bm{a_t}, \bm{s_{t+1}})$ in two steps:
\begin{enumerate}
\item Estimate the optimal policy output
\begin{equation}
\pi(\bm{s_{t+1}}) = \arg\max_{\bm{a}\in A} Q(\bm{s_{t+1}}, \bm{a})
\end{equation}
\item Minimize the Bellman error
\begin{equation}
\|Q(\bm{s_t}, \bm{a_t}) - [R(\bm{s_t}, \bm{a_t}) + \gamma \cdot Q(\bm{s_{t+1}}, \pi(\bm{s_{t+1}}))] \|
\end{equation}
\end{enumerate}

\begin{figure}
\centering
\includegraphics[width=1.0\linewidth]{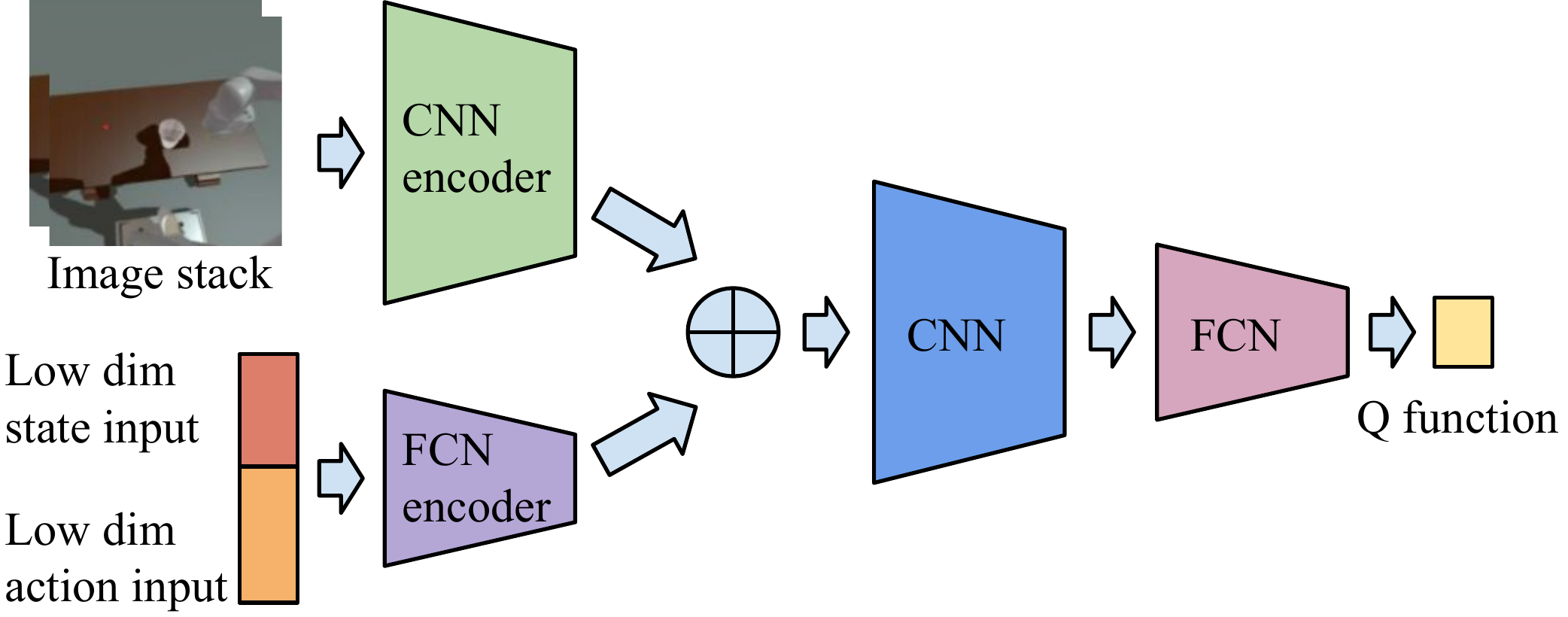}
\caption{The feed forward neural network Q function for object pushing. The stacked  current  and  initial images are fed into a convolutional neural network (CNN) encoder, and the low dimensional input is fed into a fully connected network (FCN) encoder. The output of these two streams are added together and then fed into another CNN-followed-by-FCN structure, the Q value prediction head.}
\label{fig:feedforward}
\end{figure}

For the object pushing task, we represent the Q function with a two stream deep neural network $Q_{\theta}(\bm{s_t}, \bm{a_t})$ parameterized by $\theta$, as shown in Fig. \ref{fig:feedforward}, similiar to \cite{kalashnikov2018qt}. The high dimensional stream feeds stacked current and initial images into a convolutional neural network (CNN) encoder. The low dimensional stream feeds stacked low dimensional state (gripper height and open/close status) and action (designated gripper pose, the gripper open/close command, and a termination boolean) into a fully connected network (FCN) encoder. The outputs of the two streams are added together and fed into another CNN-followed-by-FCN structure to predict the Q function value for the pushing task.

For Q function network training, we adopt QT-Opt, a distributional variant of the Q learning framework for continuous state-action tasks \cite{kalashnikov2018qt, bodnar2019quantile}. Concretely, we use distributed workers to collect data tuples $(\bm{s_t}, \bm{a_t}, \bm{s_{t+1}})$, and we store them into a replay buffer. Each optimization iteration samples a batch of data tuples. For equation (5), the optimal policy output is estimated using an online sampling-based cross entropy method (CEM) based on the current Q function. For equation (6), the Q function network weights are updated using gradient descent with the following loss function:
\begin{equation}
l(\theta) = \|Q_{\theta}(\bm{s_t}, \bm{a_t}) - [R(\bm{s_t}, \bm{a_t}) + \gamma \cdot Q_{\theta}(\bm{s_{t+1}}, \pi(\bm{s_{t+1}}))] \|
\end{equation}

\section{Contact-aware Online Context Inference}
\subsection{Online Context Inference}

The architecture in Fig. 2 has been demonstrated on challenging tasks like grasping \cite{kalashnikov2018qt}. However, given it only has access to a single sensory input, it is not able to infer the dynamics properties of the object, which is necessary for our non-planar pushing task. In this section, we describe online COntext Inference (COI), a module that takes history observation samples and encodes them into a dynamics context representation -- thus equipping the control policy with the ability to infer dynamics of the object.

\begin{figure}
\centering
\includegraphics[width=1.0\linewidth]{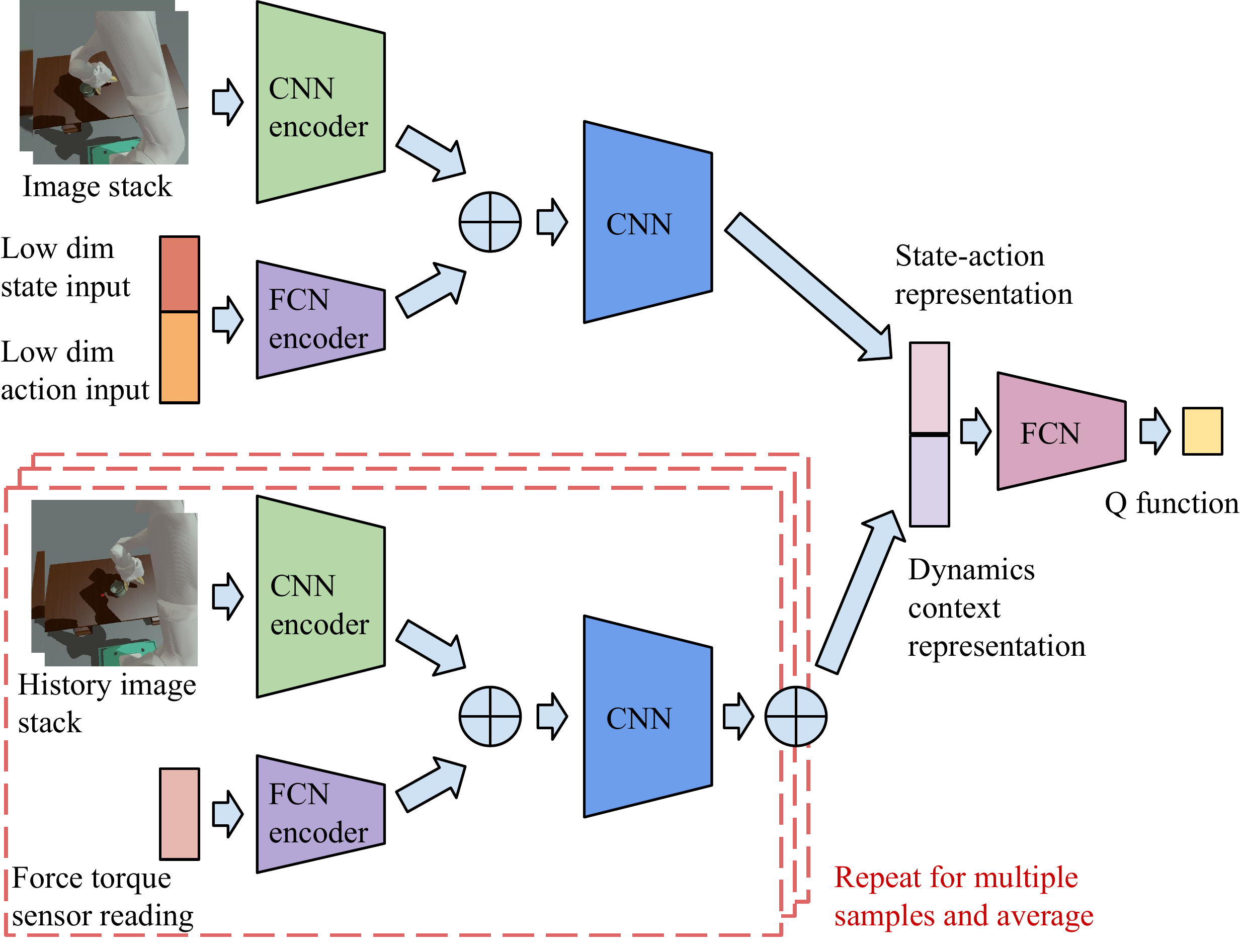}
\caption{
The proposed COntext Inference (COI) module, which works in parallel with the state-action stream. The COI takes as input a history sample consisting of the stacked pre-impact and post-impact images and the force reading, to infer the dynamics representations. The representations of different samples are averaged and concatenated to derive the state-action representation, which is then fed into a final Q value prediction network. Networks with same color share architectures, but not necessarily network weights.}
\label{fig:OCO}
\end{figure}

As shown in Figure \ref{fig:OCO}, COI consists of a set of additional streams in the policy network that encode history sensor observations into a dynamics context representation. Each stream of COI takes a pair of consecutive sensory inputs separated in time by $0.5$s (the sensor update interval in our robot system). We denote each sensory input pair as a tuple $H_{\tau} = (I_{\tau}, I_{\tau+1}, f_{\tau})$, where $I$ and $f$ refer to the camera image and force reading respectively, and $\tau$ represents the time at which the sensor inputs are retrieved. The encoded sensory input for each stream is then averaged to obtain the final dynamics context representation of COI, which is concatenated with the state-action representation to estimate the Q value.




\subsection{The Contact-aware Sampling Strategy}
\label{CO}

Given an architecture that can process multiple history sensor observation pairs, key questions are:

\begin{enumerate}
    \item How many history samples should we include?
    \item At what time should we retrieve the sensor observations? 
\end{enumerate}

With a higher number of history samples, the policy has more information to infer a potentially less noisy object dynamics representation, at the cost of higher computation time and memory. In our experiments, we found three history samples to be a reasonable number to achieve good inference performance with manageable computation cost.

\begin{figure}
\centering
\includegraphics[width=1.0\linewidth]{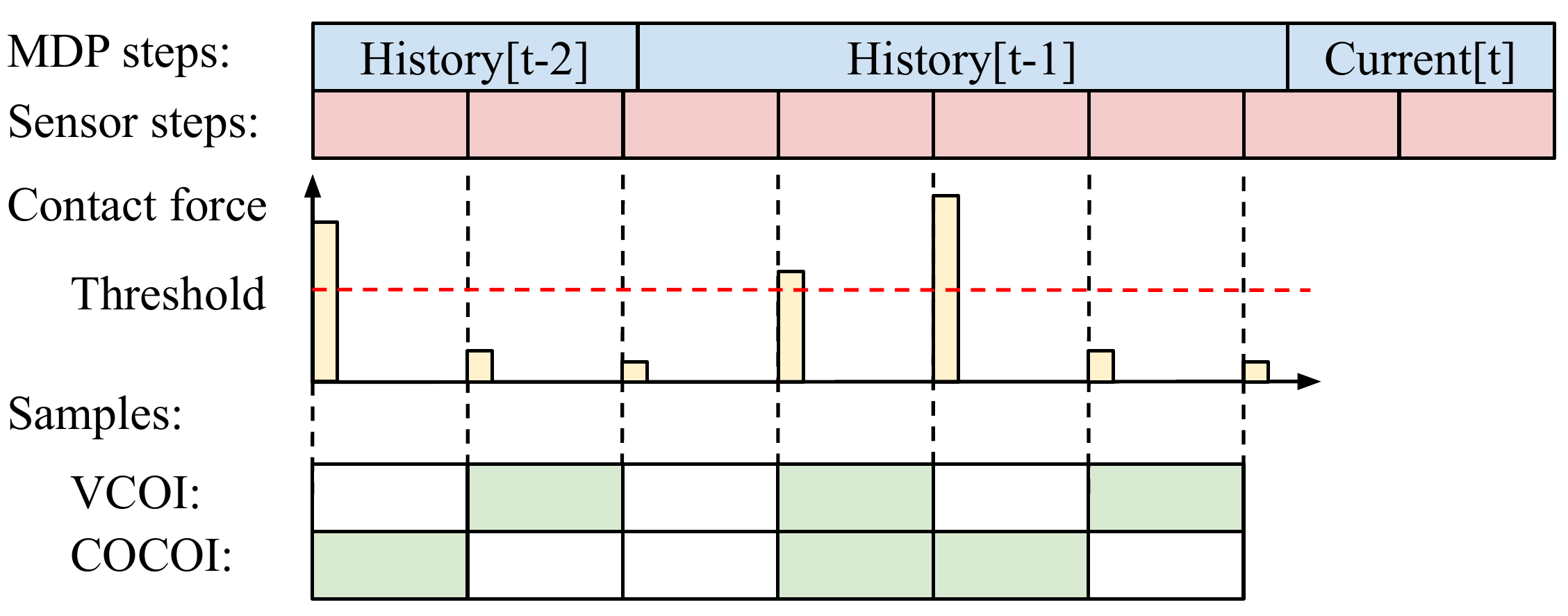}
\caption{Illustrations of the sampling strategies. For VCOI, the samples are retrieved with a uniform sampling interval. COCOI takes a contact-aware sampling method which actively checks the contact force, and only retrieve samples when the force magnitude is larger than 1 Newton.}
\label{fig:sample}
\end{figure}

The timings at which sensor observations are sampled is vital for the performance of COI. An arbitrarily sampled sensor observation pair may contain limited information about the object dynamics (e.g. when gripper is far away from the object) and contribute little to the learning performance. To ensure that each history sample contains useful information, we propose a contact-aware sampling strategy which actively checks the force torque sensor mounted at the robot gripper and only collects a sample when the contact force magnitude is considerably large ($>1$ N), as shown in Fig. \ref{fig:sample}. This strategy guarantees the samples to be representative, in that the gripper and the object are in contact. We call this sampling strategy COntact-aware-COI, or COCOI.

In our experiments, we validate the performance of COCOI by comparing it to a na\"ive strategy: vanilla COI, or VCOI, where history samples are retrieved with a uniform sampling interval, as shown in Figure \ref{fig:sample}.


\section{GAN for Visual Gap Bridging}
\label{sec:cgan}
\begin{figure}
\centering
\includegraphics[width=0.9\linewidth]{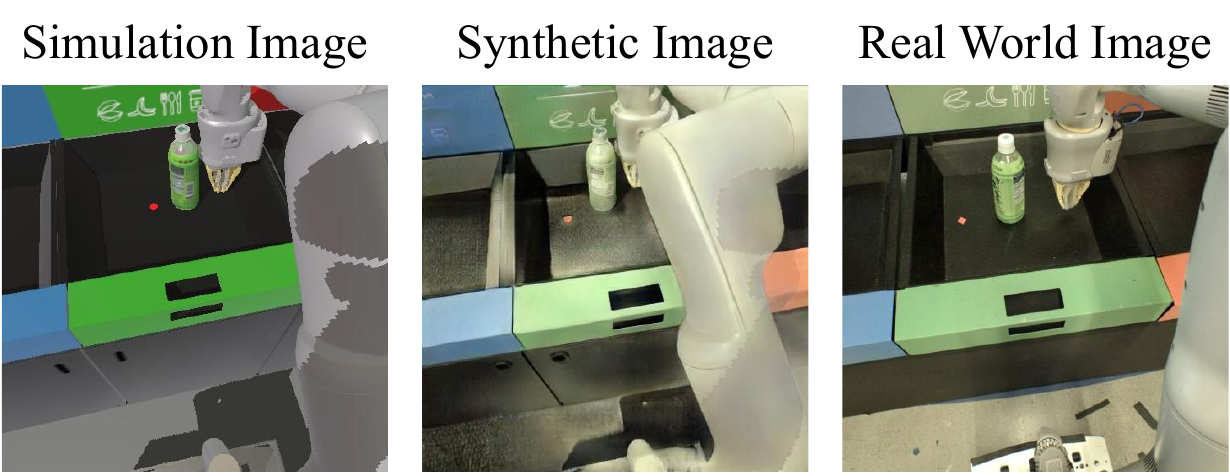}
\caption{Images of the pushing-in-station setting in simulation, with RetinaGAN visual adaptation, and in the real world.}
\label{fig:RetinaGAN}
\end{figure}

In order to adapt the RL policy trained in simulation to the real world, we also need to overcome the discrepancy between the rendered, simulated image and the image captured by a real-camera. We adopt RetinaGAN, a generative adversarial network (GAN) approach to generate synthetic images that look realistic with object-detection consistency \cite{retinagan}. See the concurrent submission\footnote{\url{https://retinagan.github.io}} for further details; this work focuses on contact inference independently of the visual discrepancy. Qualitative performance is shown in Fig. \ref{fig:RetinaGAN}. We train the RL policy with simulation data only and directly deploy it on the real robot.

\section{Experiments}
\zhuoxu{don't forget to mention: (1) the robot make smart move using learning based policy, such as open the gripper and uses the finger to push (2)}
\subsection{Setup}

\begin{figure}
\centering
\includegraphics[width=1.\linewidth]{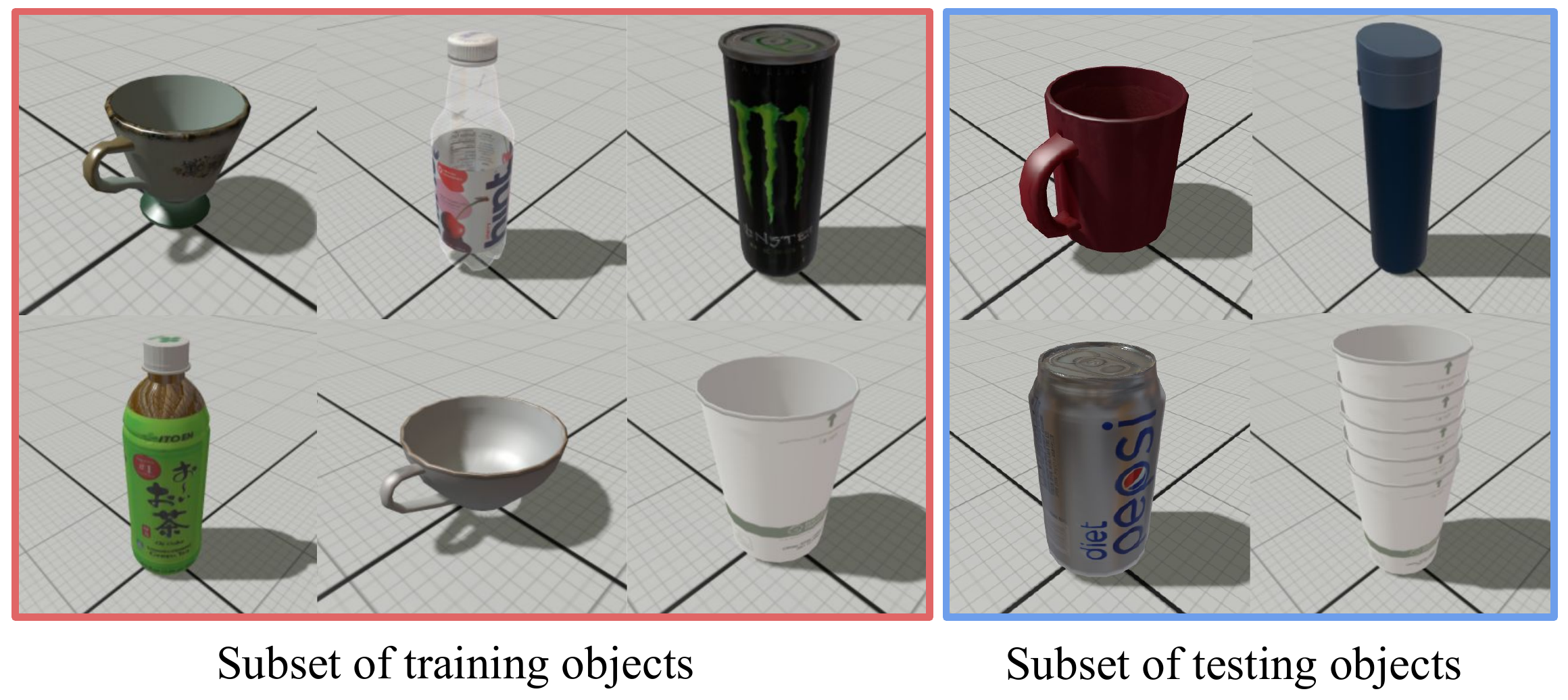}
\caption{A subset of the 75 different 3D objects used for training and testing.}
\label{fig:objects}
\end{figure}

We train the control policies in PyBullet simulation \cite{coumans2016pybullet}. We first define a flat surface randomly placed in front of the robot. The surface can be either a desk surface or a designated flat area inside a trash bin, as shown in Fig. \ref{fig:overview}. We use a 3D model set containing 75 different objects, such as cups, bottles, cans, mugs, etc. (Fig. \ref{fig:objects}). We divide the objects into a training set containing 64 objects and an unseen testing set of 11 objects including a stack of cups. Note that the upper cups in the stack have the degrees of freedom to tilt when pushed, which makes the pushing task more challenging. In PyBullet, the contact physics between the robot gripper and the object is modelled using a point contact model with an elliptic friction cone.

At the beginning of each episode, the object and the target position are randomly placed within a rectangular area. We set the object upright on the surface, and the target position is rendered using a red dot. The robot gripper is initialized at a randomized position beside the object. The push policy controls the robot gripper to push the object to the red dot. An episode is considered successful if the object is pushed to within 5 cm of the target, and the object tilting angle remains smaller than 0.1 rad. We also apply a small penalty at each timestep to encourage faster execution. The discount factor $\gamma$ of the POMDP is 0.9.

\subsection{Policy Models}
\label{sec:policy_models}

The deep neural networks for the Q functions are constructed as in Fig. \ref{fig:feedforward} and Fig. \ref{fig:OCO}, and the blocks with the same names share the same network architecture. The detailed architecture is shown in Table I. 

\begin{table}[thpb]
\centering
\caption{Architecture of each module in the Q networks}
\label{rl_module_table}
\begin{center}
\begin{tabular}{cc}
\toprule
\textbf{Block name} & \textbf{Architecture} \\ \midrule
Images input & Tensor with shape (472, 472, 6) \\ 
Low dim state input & 2 dim vector\\ 
Low dim action input & 7 dim vector\\ 
Force torque sensor reading & 3 dim vector\\ \midrule
 & Conv(64, 6, 2) \\
CNN encoder & MaxPool(3) \\
 & Repeat x6: Conv(64, 5, 1) \\
 & MaxPool(3) \\ \midrule
 & FC(256) \\
FCN encoder & FC(64) \\
 & Reshape(1, 1, 64) \\ \midrule
 & Conv(64, 3, 1)\\
CNN & MaxPool(2)\\
 & Repeat x3: Conv(64, 3, 1)\\ \midrule
state-action representation & (8, 8, 64) dim matrix\\ 
dynamics context representation & (8, 8, 64) dim matrix \\ \midrule
 & FC(64)\\
FCN & FC(64)\\
 & Sigmoid \\ \midrule
Q function & 1 dim vector \\\bottomrule
\end{tabular}
\end{center}
\end{table}

\begin{table*}[thpb]
\centering
\caption{Comparison of success rate for models evaluated with different initial placement settings.}
\begin{center}
\begin{tabular}{cccccccc} 
\toprule
\textbf{Initial range} & \textbf{0.3m x 0.6m} & \textbf{0.3m x 0.5m} & \textbf{0.3m x 0.4m} & \textbf{0.3m x 0.3m} & \textbf{0.25m x 0.5m} & \textbf{0.25m x 0.5m}\\ \midrule
Baseline & 26.5\% & 34.1\% & 51.1\% & 59.2\% & 40.0\% & 42.5\% \\ 
COI & 36.0\% & 43.2\% & 63.4\% & 72.2\% & 49.0\% & 50.8\%\\ 
Oracle & \textbf{43.8\%} & 53.9\% & \textbf{73.8\%} & \textbf{78.2\%} & 60.5\% & 62.4\%\\ 
COCOI & 43.0\% & \textbf{59.3\%} & 73.2\% & 75.7\% & \textbf{64.7\%} & \textbf{62.6\%}\\
\bottomrule
\end{tabular}
\end{center}
\label{tab:distance}
\end{table*}

\begin{table*}[thpb]
\centering
\caption{Comparison of success rate for models evaluated with different dynamics properties settings.}
\label{tab:dynamics}
\begin{center}
\begin{tabular}{ccccccc}
\toprule
\textbf{Model} & \textbf{default setting} & \textbf{0.0-0.5 friction} & \textbf{1.0-1.5 friction} & \textbf{0.5-1.0kg mass} & \textbf{Unseen objects} & \textbf{Cup stack}\\ \midrule
Baseline & 51.1\% & 53.8\% & 32.1\% & 32.6\% & 42.5\% & 26.2\% \\ 
COI & 63.4\% & 59.6\% & 44.2\% & 44.7\% & 48.3\% & 38.8\% \\ 
Oracle & \textbf{73.8\%} & 39.6\% & 38.2\% & 41.4\% & \textbf{60.9\%}& 36.5\% \\ 
COCOI & 73.2\% & \textbf{72.6\%} & \textbf{47.2\%} & \textbf{49.9\%} & 58.9\% & \textbf{43.7\%} \\ \bottomrule
\end{tabular}
\end{center}
\end{table*}

We train and compare four models:

\begin{enumerate}
    \item The baseline model: the basic feed forward Q function network as described in Section \ref{baseline} and shown in Fig. \ref{fig:feedforward}. This model is the most straightforward and shows the capability of the baseline RL method.
    \item The VCOI model: the VCOI method as shown in Figure \ref{fig:OCO}. The history observations are sampled using uniform sampling.
    \item The oracle: the architecture of this model is the same as the baseline, but we expand the low dimensional state input with 2 key, unobservable dynamics parameters: the object mass and friction coefficient. We directly extract the ground truth parameter values from the simulator to obtain an oracle model.
    \item The COCOI: the proposed COntact-aware Online COntext inference model as shown in Figure \ref{fig:OCO}. The active contact-aware sampling strategy is applied.
\end{enumerate}

\subsection{Policy Learning and Comparison}

We adopt the QT-Opt framework \cite{kalashnikov2018qt} to train the policies. In the training setting, the objects and goal locations are randomly sampled in a 0.5m $\times$ 0.3m area, the object mass is randomized from 0.05 kg to 0.5 kg, and the friction coefficient is randomized from 0.5 to 1.0. We train the models using stochastic gradient descent, using a learning rate of 0.0001 and momentum of 0.9. We train the models with a batch size of 2048 for 80k steps.

\begin{figure}
\centering
\includegraphics[width=0.9\linewidth]{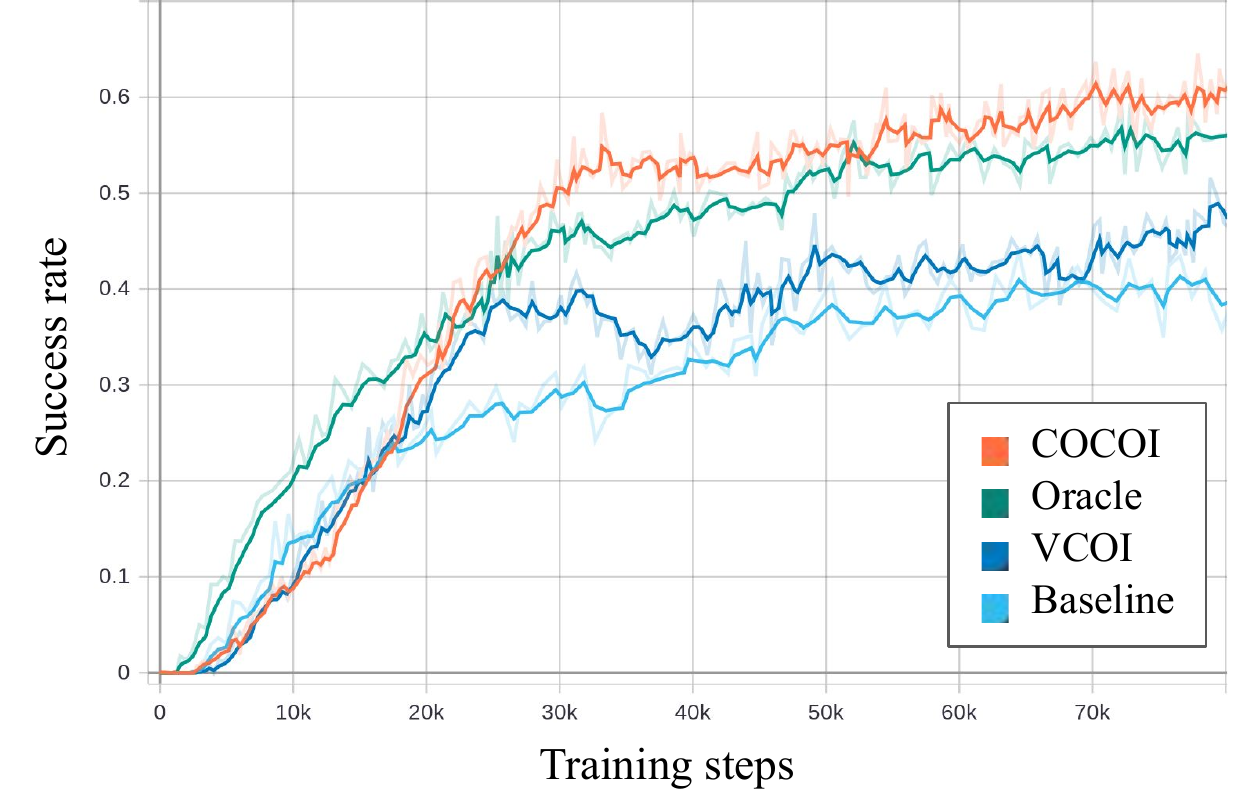}
\caption{Training success rate as a function of training steps for the four RL models in Section \ref{sec:policy_models}}.
\label{fig:training_log}
\end{figure}

In the early stage of training, we use a rule-based scripted policy, which moves the gripper along the line connects the object and the target, to generate successful episodes and improve the exploration efficiency. This rule-based method only achieves less than 5\% success rate, illustrating the difficulty of the non-planar pushing task. During training, we observe that the policy first obtains the capability to solve easier scenarios where the object-goal distance is short and then gradually learns to push objects that are initialized far from the goal. Fig. \ref{fig:training_log} shows a comparison of training performance for the four models.

As shown in Fig. \ref{fig:training_log}, COCOI and the oracle model perform significantly better than VCOI and the baseline model. COCOI reaches even higher success rate than the oracle, which indicates that COCOI is capable to capture more than just the oracle information (object mass and friction coefficient) - there are other factors such as the object shape and contact point that affect the dynamics properties.

\subsection{Performance under Domain Variations}

Overall, the RL policy succeeds in pushing the upright object and learns to perform smart behaviors. For example, the robot can break contact with object when the object leans, and it can open its gripper to use its finger to make subtle impacts. We evaluate pushing performance for the four models on a large variety of settings.

First, the initial range of the object and the target are varied and the results are shown in Table \ref{tab:distance}. COCOI consistently outperforms VCOI and baseline and achieves a similar performance to the oracle. Specifically, COCOI shows an average relative improvement of 50\% and 20\% success rate compared to the baseline and VCOI, respectively.

Second, we fix the initial object placement to 0.4m $\times$ 0.3m and vary dynamics properties. We change the friction coefficient and the object mass to be outside the training range. We also test performance of the models on unseen objects and a stack of cups whose inertia can change during pushing. Evaluation results with these setting variations are reported in Table \ref{tab:dynamics}.

Across different domains, COCOI consistently outperforms other methods. The performance of the oracle model is especially poor in cases where the real friction parameter is not in the range of the training set. This could be due to the policy overfitting to the input dynamics parameters.

\subsection{Interpretation of Context Representations}

\begin{figure}
\centering
\includegraphics[width=0.8\linewidth]{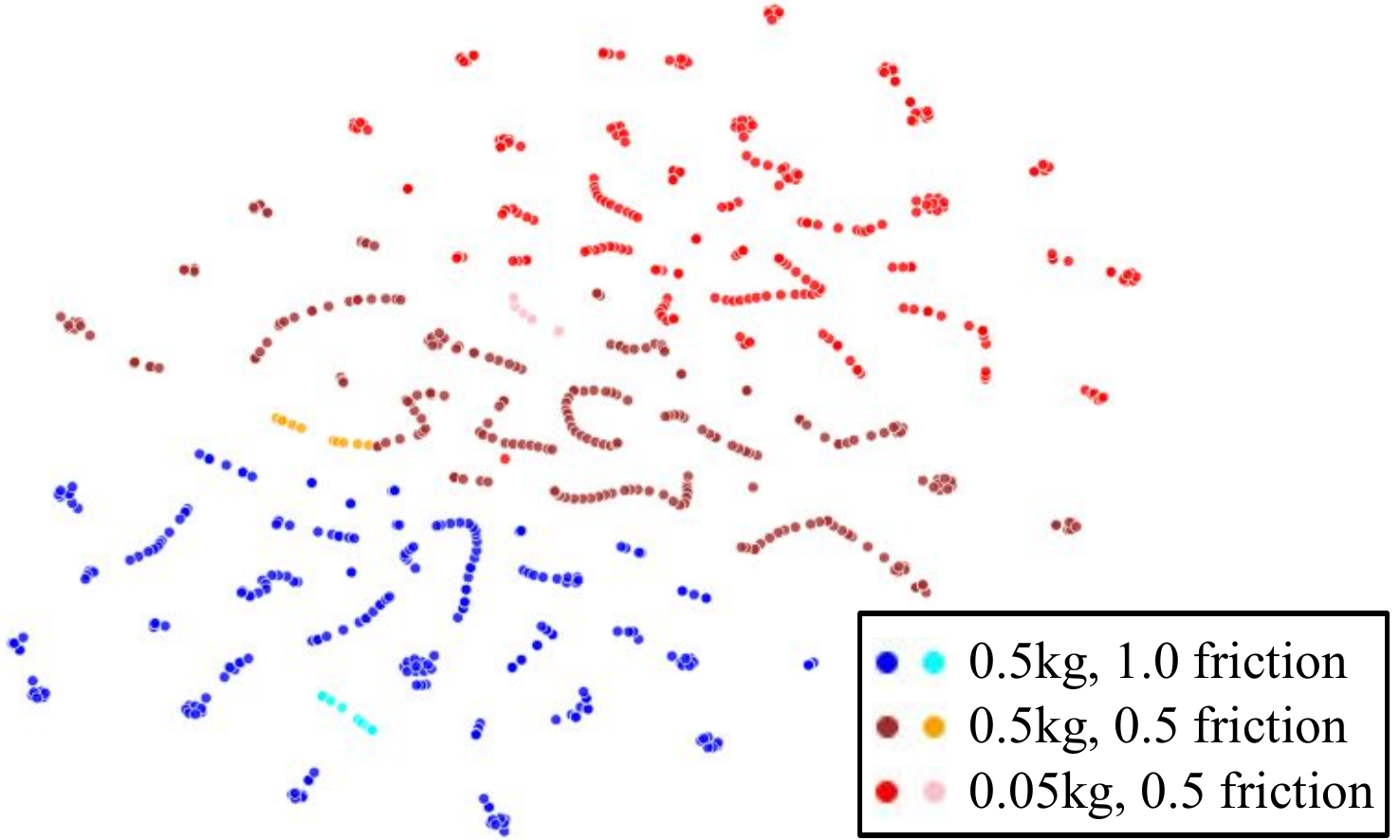}
\caption{t-SNE visulization of the inferred context representation for three different dynamics parameters settings. For each setting, the context representations from a randomly chosen episode is highlighted with a brighter color. The clusters show clear separation and are distributed with an order consistent with the friction magnitude.}
\label{fig:representation_clusters}
\end{figure}

To inspect the dynamics context learned by COCOI, we visualize the inferred representations for three settings with different dynamics parameters. For each setting, we run our controller for 20-30 episodes to fill a buffer of 256 dynamics context representations. We then visualize these representations using a combination of principle component analysis (PCA) and t-distributed stochastic neighbor embedding (t-SNE) \cite{JMLR:v9:vandermaaten08a} (Fig. \ref{fig:representation_clusters}). The visualization shows clear separation between settings, which indicates COCOI learns to infer the dynamics properties. Also, representations within one episode are grouped closer to each other than to other episodes, suggesting that learned representations are consistent and structured. Moreover, we observe the order of the clusters is consistent with the dynamics properties: the clusters with the largest mass and friction and the smallest mass and friction are farthest apart, while the cluster with intermediate parameters is in the center.

\subsection{Real World Deployment}

\begin{figure}[th]
\centering
\includegraphics[width=0.8\linewidth]{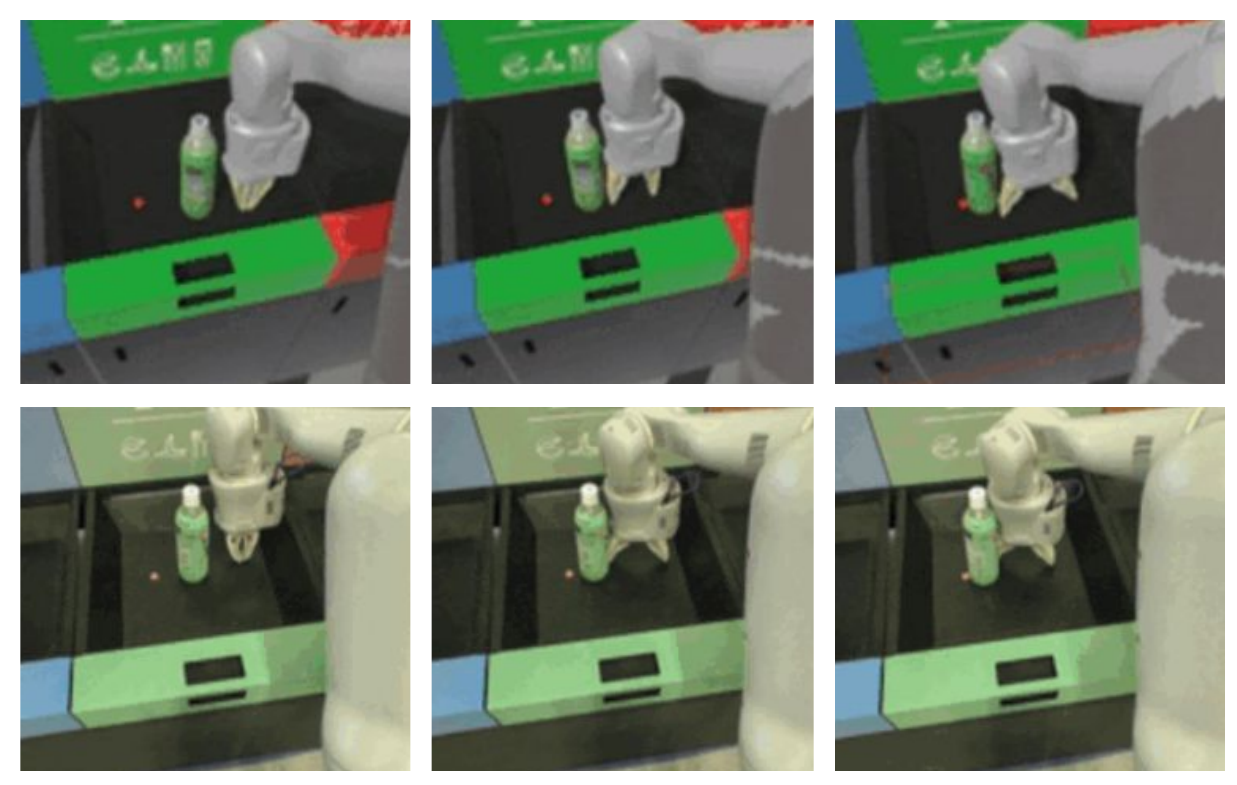}
\caption{Visualization of sim-to-real pushing policy transfer.}
\label{fig:sim2real_episodes}
\end{figure}

To test real world deployment, we design a push-in-bin task in both the simulator and in the real world, as shown in Fig. \ref{fig:overview}. We adopt the method described in Section \ref{sec:cgan} and train a RetinaGAN model to adapt the simulation images to synthetic images with realistic appearance. We train the pushing policy with COCOI based on synthetic images and run 10 real world pushing episodes. We achieve $90\%$ success, demonstrating the capability of our 3D pushing policy to overcome both the visual and dynamics domain gap. Fig. \ref{fig:sim2real_episodes} shows example sequences in simulation and the real world.

\section{Conclusions}

We propose COCOI, a deep RL method that uses history robot-object interaction samples to infer contact dynamics context, and we show it outperforms baseline in contact-rich manipulation tasks with domain variations. We design and study COCOI on a non-planar pushing task commonly seen in everyday life. Extensive experiments demonstrate the capability of COCOI in a wide range of settings, dynamics properties, and sim-to-real transfer scenarios.

There are many promising future work directions to pursue based on our approach. For example, we study the non-planar pushing task with a single object on the surface. It would be interesting to train manipulation controllers that can perform pushing with multiple objects or in a cluttered environment. In addition, extending our approach to push non-rigid objects such as a piece of cloth is another important direction that can further expand the capability of our controller.

\bibliography{references}
\bibliographystyle{ieeetr}
\end{document}